\documentclass[letterpaper, 10 pt, conference]{ieeeconf}  

\IEEEoverridecommandlockouts                              

\usepackage{graphicx} 
\usepackage{stfloats}
\usepackage{epsfig} 
\usepackage{hyperref}
\usepackage{amsmath}
\usepackage{booktabs}
\usepackage{xcolor}

\title{\LARGE \bf

Threading Optimization for Vision-Language-Action Model Inference in Low-Cost Smart Agricultural Manipulation 

}

\begin{document}


\author{
Keith Truongcao$^{*}$,
Christopher Nhu$^{*}$, 
Zijian An$^{}$,
Phong Nguyen$^{}$,
Siwei Cai$^{}$,
and Lifeng Zhou$^{\dagger}$
\thanks{$^{*}$ Equal contribution} %
\thanks{$\dagger$ Corresponding author}%
\thanks{Department of Electrical Engineering, Drexel University, Philadelphia, USA}%
}

\maketitle
\thispagestyle{empty}
\pagestyle{empty}

\begin{abstract}
Vision-Language Action (VLA) models continue to face challenges such as slow inference speed and difficulty performing fine-grained motion adjustments, limiting their widespread adoption in industry. While the Real-Time Action Chunking (RTAC) algorithm has been proposed to address these bottlenecks, bridging the gap between the algorithm provided in pseudocode to a stable, real-world deployment on a low-cost robotic arm remains a challenge. In this work, we present a complete system-level implementation of RTAC tailored for a low-cost robotic manipulation system. We advance beyond the original high-level pseudocode by optimizing the threading implementation for the policy inference and control pipeline, reducing end-to-end latency and improving responsiveness without modifying the underlying policy. We evaluate this system on tasks involving the manipulation of agricultural produce, specifically garlic bulbs and walnuts. Experimental results demonstrate that our custom threading implementation significantly improves control stability and speed compared to the base implementation of RTAC. \href{https://youtu.be/Ryvi5j6MjXY}{A video} of our paper is available online. 
\end{abstract}

\section{Introduction}

Recent advancements in robotic learning have been driven by the emergence of Vision-Language Action (VLA) models, which leverage large-scale pre-training to enable robots to understand natural language commands to generate actions to perform a variety of tasks. Models such as OpenVLA \cite{kim2024openvla} and $\pi_0$ \cite{black2024pi_0} have demonstrated impressive semantic understanding, and a wide range of downstream applications built upon these models have since been explored \cite{an2025claw,yang2025seqvla}. However, their deployment in dynamic, real-world environments remains bottlenecked by inference latency. In addition to that, large transformer-based policies often operate at lower frequencies than typical control frequencies used by robotic hardware \cite{yu2025survey}, causing an increased amount of jitter, and an inability to perform fine-grained corrections.

To mitigate these issues, temporal ensemble methods such as Real Time Action Chunking (RTAC)~\cite{black2025real} have been proposed. RTAC theoretically allows for smooth control by generating action chunks via inpainting that are executed asynchronously while the next inference step is computed, essentially giving it the ability to think while moving. However, while the original proposal for RTAC provides the mathematical foundation and high-level pseudocode implementation, it abstracts away the complex systems-level challenges required to implement itself on physical hardware. Bridging the gap between a theoretical threading model and a deterministic, real-time controller is a non-trivial engineering challenge. Naive threading implementations based on the provided pseudocode suffer from not being able to achieve true parallelism, race conditions, and thread locking, which can reintroduce the jitter and latency that RTAC was designed to remove. 

\begin{figure*}
    \centering
    \includegraphics[width=\textwidth]{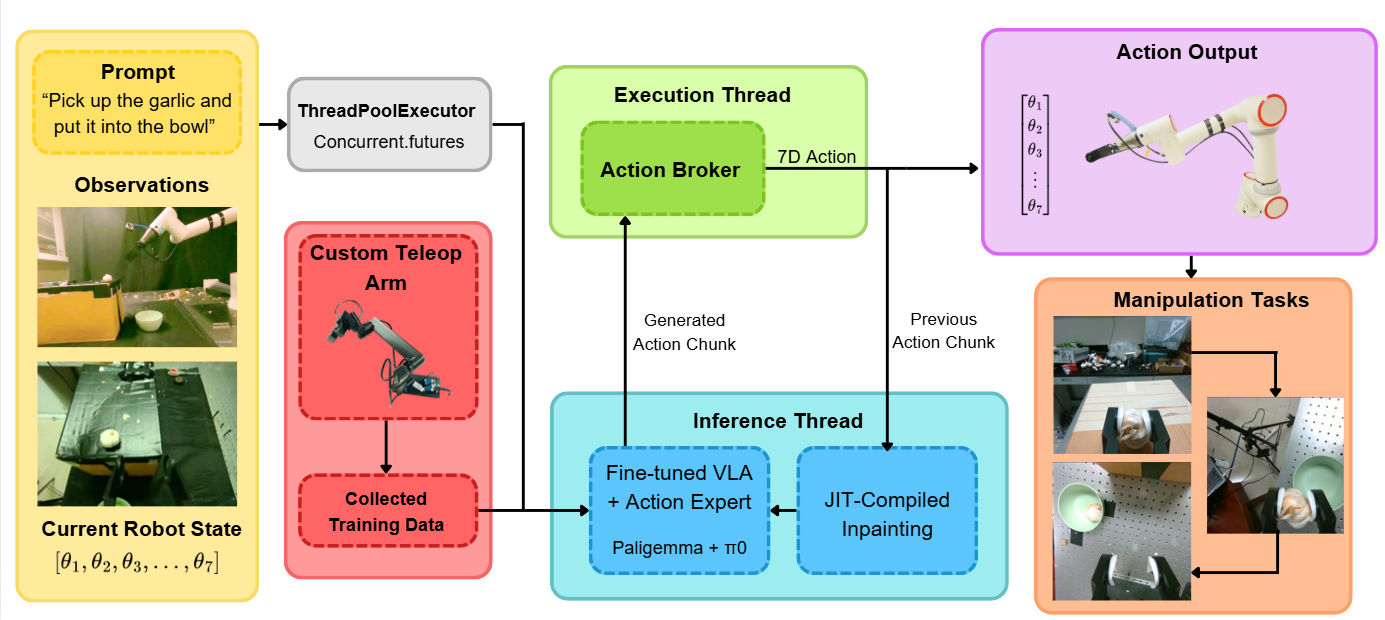}
    \caption{Given the language prompt ``Pick up the garlic and put it into the bowl," two camera observations, and the robot's 7D joint state, the ThreadPoolExecutor sends the input to the Inference Thread. This thread runs a PaliGemma LoRA (fine-tuned on data from our custom 3D-printed GELLO teleoperation arm that mirrors the FR5) paired with the $\pi_{0}$ Action Expert to generate action chunks. The action chunks are sent to the Action Broker in the Execution Thread, which feeds actions to the FR5 robotic arm. The tail end of the current chunk (the previous chunk to the future generated chunk) feeds back to the Inference Thread to be used in our JIT-compiled inpainting to asynchronously generate the next action chunk that smoothly continues the robot's motion. } 
    \label{fig:pipeline}
\end{figure*}

In this work, we present a robust, system-level implementation of RTAC tailored for a low-cost robotic system ($<$\$6,000) based on the Fairino FR5, a low-cost industrial robotic arm ($<$\$4,000). We move beyond the high-level abstractions of the original proposal to architect a custom threading implementation, as detailed in Figure~\ref{fig:pipeline}, that explicitly manages the asynchronous communication between the high-latency VLA policy and the high-frequency robot controller. By optimizing the producer-consumer relationship between inference and execution, we achieve a system that is faster overall in comparison to the base implementation of RTAC.

The primary contributions of this paper are as follows:
\begin{itemize}
\item We introduce an accessible, low-cost hardware platform that includes the FR5 manipulator, dual camera setup, and a teleoperation interface utilizing the open-source GELLO framework alongside 3D-printed hardware \cite{wu2024gello}.
\item We propose an improved threading optimization that decouples policy inference from actuation on the Fairino FR5, translating the RTAC pseudocode into a stable control system architecture that minimizes end-to-end latency.
\item We demonstrate that our improved implementation enables the manipulation of agricultural produce (specifically garlic bulbs and walnuts) faster than baseline control schemes, which validates the stability of our approach.
\end{itemize}
While prior research has validated the performance of RTAC
within VLA pipelines~\cite{black2025real}, current RTAC literature ignores low-level robotic deployment challenges. In particular, the Python Global Interpreter Lock (GIL) as noted by Mitchell et al. \cite{lim2023python}, enforces serialized execution, which can impede real-time responsiveness.


Our pipeline, as shown in Figure~\ref{fig:pipeline}, feeds inputs such as a language prompt, dual camera observations, and robot state into a inference thread containing a Paligemma LoRA VLA and $\pi_0$ action expert for action chunk generation. These action chunks are sent to an execution thread with an Action Broker that supplies individual 7D actions to the robot arm.

To the best of our knowledge, this work represents the first dedicated exploration of the systems-level challenges associated with integrating RTAC into a physical VLA pipeline. Unlike previous studies that focus primarily on policy training or algorithmic modifications, we explicitly address the engineering gap between abstract pseudocode and stable execution. Furthermore, we are the first to propose and validate a specific optimization of the baseline threading architecture, effectively simplifying the integration process for future researchers while simultaneously enhancing control authority for precision tasks.

The remainder of this paper is organized as follows: Section II reviews related work in VLA control and action chunking; Section III delves into our methodology; Section IV presents our approach, including our experimental setup; Section V shows our experimental results graphically and numerically; Section VI concludes with directions for future work.

\section{Related work} 
\subsection{Vision-Language Action Models} 
Recent advances in large language models (LLMs) have catalyzed the development of VLA models, which ground semantic reasoning in physical control. Models such as RT-2 \cite{brohan2023rt2} and PaLM-E \cite{driess2023palme} demonstrated that pre-trained vision-language backbones could be fine-tuned to output robotic actions directly. More recently, OpenVLA \cite{kim2024openvla}, $\pi_0$ \cite{black2024pi_0} $\pi_{0.5}$, \cite{physicalintelligence2025pi05} $\pi_{0.6}$,   \cite{physicalintelligence2025pi06}, and, $\pi_{0.6}*$\cite{amin2025pistar06} have democratized access to these policies by providing open-source weights and efficient architectures. However, while these models exhibit impressive generalization capabilities, their inference latency often precludes high-frequency control. Most VLA architectures, based on large transformer backbones, operate in the 3–10 Hz range \cite{kim2024openvla}, creating a significant bandwidth mismatch with industrial robot controllers that require hard real-time inputs at 500 Hz or above.

\subsection{Temporal Action Chunking}
To address the latency bottleneck, temporal ensembling techniques such as Action Chunking have emerged as a dominant paradigm. The Action Chunking Transformer (ACT) \cite{zhao2023act} introduced the concept of predicting a sequence of future actions in a single forward pass, allowing the robot to execute smooth trajectories while the next inference step computes. This approach was further formalized in RTAC \cite{black2025real}, which provides a theoretical framework for asynchronous execution. While these works provide the algorithmic foundation, they often abstract away the system-level implementation details required to deploy these policies on cost-constrained hardware. Our work bridges this gap by providing a concrete, open-source implementation of the RTAC runtime tailored for standard industrial manipulators.

\subsection{Real-Time Systems in Robotics}
The challenge of integrating non-deterministic perception with deterministic control is a classical problem in robotics. Traditional approaches utilize Real-Time Operating Systems (RTOS) or dedicated hardware bridges to ensure safety and timing guarantees \cite{siciliano2009}. However, the modern shift toward Python-centric deep learning pipelines has reintroduced significant latency and jitter issues. As noted by Mitchell et al. \cite{lim2023python}, the Python GIL serializes execution threads, creating a ``saturation cliff" where increased threading for I/O (like camera streaming) paradoxically degrades real-time performance. While prior works have addressed this via multi-process architectures or C++ bindings \cite{murali2019pyrobot}, few have specifically optimized the JAX-based \cite{frostig2018jax} inference pipeline used by modern VLAs like $\pi_0$. Our work introduces a specialized producer-consumer architecture that specifically targets the latency issues in Python-based VLA deployment.
\section{Preliminaries}

\subsection{Vision-Language Action Models}

Recent VLA models \cite{kim2024openvla, black2024pi_0, driess2023palme, physicalintelligence2025pi05, physicalintelligence2025pi06, amin2025pistar06} have shifted from discretized tokens to predicting continuous action chunks $A_t = [a_t, \dots, a_{t+H-1}]$ over horizon $H$ from multimodal observations $o_t$. $\pi_0$~\cite{black2024pi_0} frames action generation as a continuous-time probability flow by learning a velocity field $v_\theta(A_t^\tau, o_t)$ to map Gaussian noise to action trajectories. During inference, an ODE solver (e.g., forward Euler with step $\delta$) synthesizes continuous control actions via iterative updates:
\begin{equation}
    A_t^{\tau+\delta} = A_t^\tau + \delta v_\theta(A_t^\tau, o_t).
\end{equation}

\subsection{Real-Time Action Chunking}

Traditional infer-and-then-act paradigms introduce latency, causing jerky robotic movements. RTAC~\cite{black2025real} mitigates this using an asynchronous, dual-thread architecture that decouples physical execution from inference.
To ensure smooth transitions, newly inferred chunks must temporally align with the executing trajectory $\mathbf{Y}$ over overlapping timesteps masked by $\mathbf{W}$. This is framed as a guided flow-matching temporal inpainting problem using a modified velocity field:
\begin{equation}
\begin{split}
    \mathbf{v}_{\Pi\text{GDM}}(\mathbf{A}_t^\tau, \mathbf{o}_t, \tau) &= \mathbf{v}(\mathbf{A}_t^\tau, \mathbf{o}_t, \tau) \\
    &\quad + \min\left(\beta, \frac{1-\tau}{\tau \cdot r_\tau^2}\right) \left(\mathbf{Y} - \widehat{\mathbf{A}_t^1}\right)^\top \\
    &\quad \times \text{diag}(\mathbf{W}) \frac{\partial \widehat{\mathbf{A}_t^1}}{\partial \mathbf{A}_t^\tau} \label{eq:guided_inference}
\end{split} 
\end{equation}
where the approximated final chunk $\widehat{\mathbf{A}_t^1}$ is:
\begin{equation}
    \widehat{\mathbf{A}_t^1} = \mathbf{A}_t^\tau + (1-\tau)\mathbf{v}(\mathbf{A}_t^\tau, \mathbf{o}_t, \tau),
\end{equation}
and the variance scaling schedule $r_\tau^2$ is:
\begin{equation}
    r_\tau^2 = \frac{(1-\tau)^2}{\tau^2 + (1-\tau)^2}.
\end{equation}
This formulation specifically pulls $\widehat{\mathbf{A}_t^1}$ toward $\mathbf{Y}$ at masked steps. The hyperparameter $\beta$ ensures stability, and the Jacobian propagates spatial constraints, inferring the correct motions with stability and little latency.

\section{Approach}
Our approach, as seen in Figure~\ref{fig:pipeline}, aims to enable efficient RTAC for robotic manipulation by speeding up a VLA on a low-cost robotic system. We first describe the hardware platform used for data collection and VLA deployment, including the robotic manipulator, sensing setup, and teleoperation interface. We then present the system-level integration connecting the policy server, runtime controller, and robot hardware, emphasizing the asynchronous threading architecture that supports efficient action execution.
\subsection{Robotic Manipulation System} 

\begin{figure} 
    \centering
    \includegraphics[width=\columnwidth]{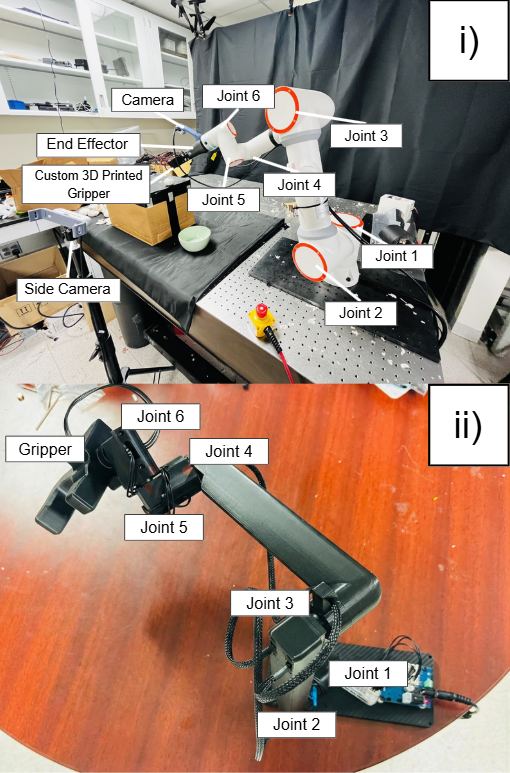}
     \caption{Labeled Depiction of the robotic platform. Figure (i) contains the Fairino FR5 cobot arm with its 6 joints labeled, the Jodell end effector equipped with a custom 3D-printed gripper, and two cameras on the wrist and side. Figure (ii) shows a custom 3D-printed teleoperated arm mirroring the FR5 cobot arm, also with six labeled joints and a gripper, used for collecting training data.}
    \label{fig:arms_setup}
\end{figure}

We developed a robotic platform, depicted in Figure~\ref{fig:arms_setup}, centered around the Fairino FR5 collaborative robotic arm, which has 6 joints and 6 degrees of freedom (DOF), can manage a 5 kg payload, and has a reach of 854 mm. As shown in Figure~\ref{fig:arms_setup}i, the FR5 robot arm's 6 total joints are labeled alongside the end effector attached at the end. The end effector is a RG52-050 Electric Gripper by Jodell Robotics, which has an adjustable stroke of 52 mm, a maximum clamping force of 60 N, and provides another degree of freedom to bring the total to 7-DOF. We designed a custom 3D-printed two-finger gripper and attached it to the Jodell Gripper to manipulate objects. Visual Observations were captured using two cameras: the one mounted onto the wrist is an Intel Realsense D405, and the stand-mounted camera is an Intel Realsense D455. 

To collect demonstration data, we integrate the FR5 manipulator with the low-cost, open-source Gello Arm teleoperation device~\cite{wu2024gello}, as shown in Figure~\ref{fig:arms_setup}ii. The Gello Arm is 3D-printed and assembled using bus servos, and its joint angles are read at a fixed control frequency. A translation layer is developed to align the motion of the Gello Arm with the FR5, filter sensor noise, and smooth command signals to suppress jitter. The FR5 operates in a closed-loop joint-space servo mode, enabling low-latency tracking and stable teleoperation, while allowing synchronized robot state and action trajectories to be recorded during demonstrations.

The robot is connected to the main system via Ethernet. The system was deployed on a workstation equipped with an NVIDIA RTX 4090, with 24GB of VRAM. The VLA policy server and the real-time control client were executed on the same machine, communicating via a local WebSocket connection.




\subsection{System Level Integration}

\begin{figure*}[!t]
    \centering
    \includegraphics[width=\textwidth]{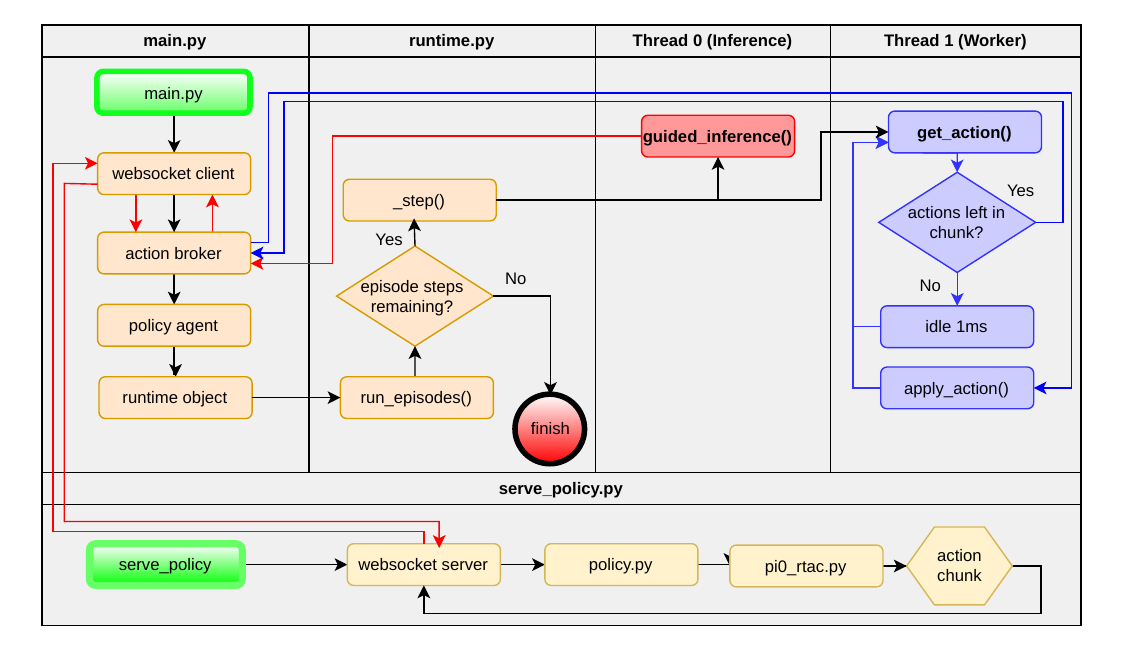}
    \caption{Dual-thread action generation architecture. \texttt{serve\_policy.py} hosts the WebSocket server, while \texttt{main.py} runs \texttt{runtime.py}. Thread 0 (inference) sends observations via WebSocket to generate action chunks with \texttt{pi0\_rtac.py}, seamlessly supplying the action broker for uninterrupted execution. Concurrently, Thread 1 (worker) executes actions supplied by the action broker.}
    \label{fig:code_architecture}
\end{figure*}

Implementing the baseline RTAC architecture \cite{black2025real} into a concrete Python execution environment initially appeared straightforward. However, practical integration revealed structural latency bottlenecks that the RTAC  pseudocode did not address.


To resolve these bottlenecks, we optimized the threading architecture. In our improved asynchronous approach  (Figure~\ref{fig:code_architecture}), Thread 0 is completely dedicated to performing guided inference as shown in Equation~\ref{eq:guided_inference}, which generates a new action chunk.  Meanwhile, Thread 1 handles the physical hardware execution, applying actions whenever a new action chunk becomes available from Thread 0 via the Action Broker, which exists to execute the action chunks from JAX \cite{frostig2018jax} into real actions on the robot.

Traditionally, real-time control loops in Python rely on the standard \texttt{threading} module, which is constrained by the GIL and prevents multiple threads from executing Python bytecode simultaneously. This restriction introduces unpredictable latency, a critical bottleneck for high-frequency policy inference and continuous robotic control. While recent updates like Python 3.13 promise true parallel multithreading by optionally disabling the GIL, physical robotic systems often necessitate backward compatibility with older, stable Python versions.

To achieve efficient concurrency without sacrificing this compatibility, our architecture leverages \texttt{concurrent.futures}, a thread pool manager library, alongside JAX-based JIT compilation for accelerated execution. Our implementation utilizes the thread-pool executor structure illustrated in Figure~\ref{fig:pipeline}, which yields noticeable performance improvements for VLA action and inference workloads. Because our control loop involves heavy I/O operations and asynchronous hardware communication, the Python interpreter naturally releases the GIL while Thread 0 waits for GPU kernel execution during guided inference. \texttt{Concurrent.futures} efficiently manages these resources under the hood, allowing the physical action execution in Thread 1 to run concurrently. This robust threading implementation not only simplifies the development of the asynchronous dual-thread architecture, but also provides backward-compatible resource scheduling that minimizes latency.

\section{experiments}
\begin{figure*}[!t]
    \centering
    \includegraphics[width=\textwidth]{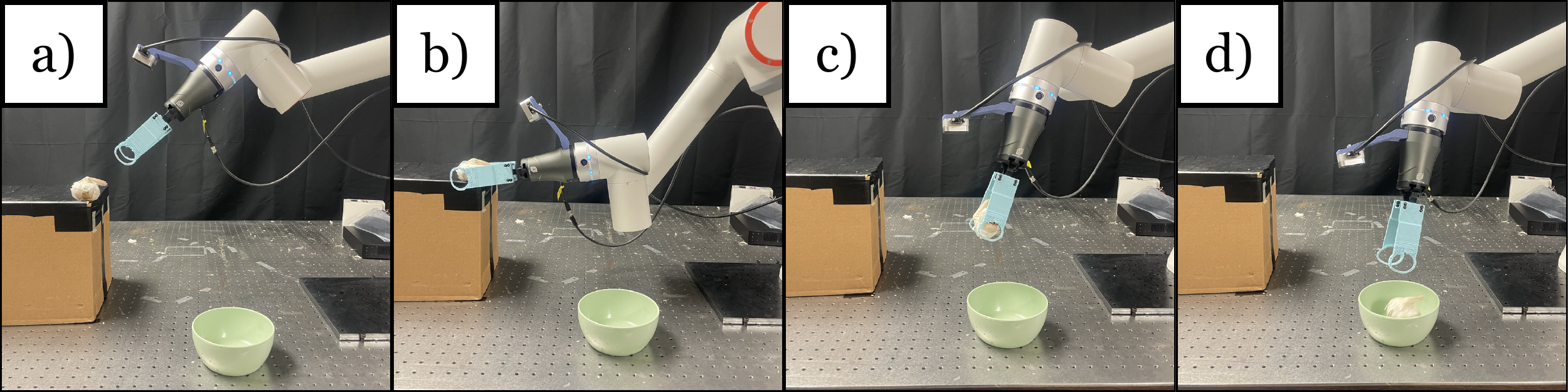}
    \caption{Task 1: Grasping a garlic at a precise position and orientation and depositing it into a bowl.}
    \label{fig:garlic_task}
\end{figure*}

\begin{figure*}[!t]
    \centering
    \includegraphics[width=\textwidth]{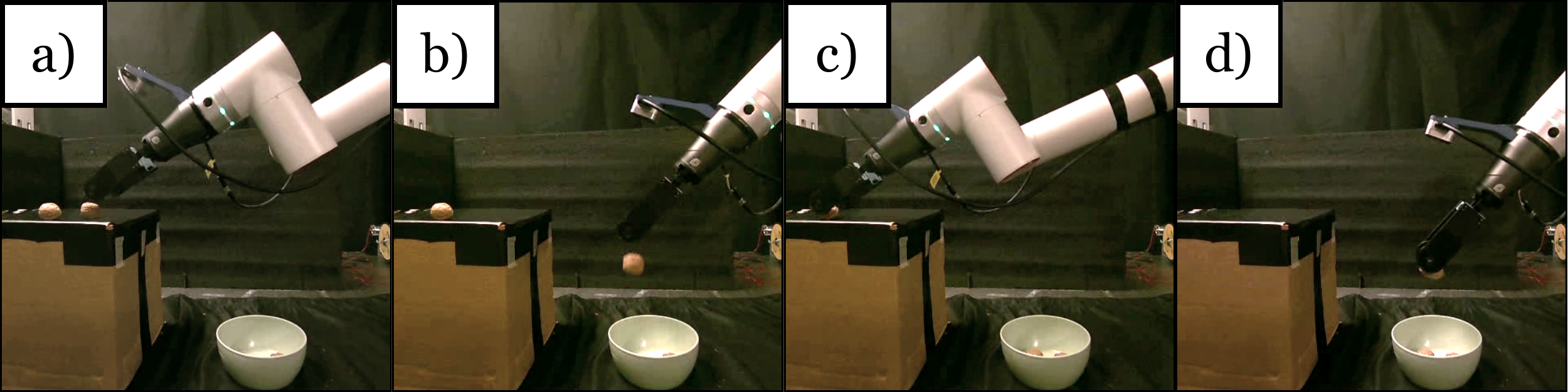}
    \caption{Task 2: Sequential grasping of two walnuts, one at a time, to deposit them into a bowl.}
    \label{fig:walnut_task}
\end{figure*}
\begin{figure*}[!t]
    \centering
    \includegraphics[width=\textwidth]{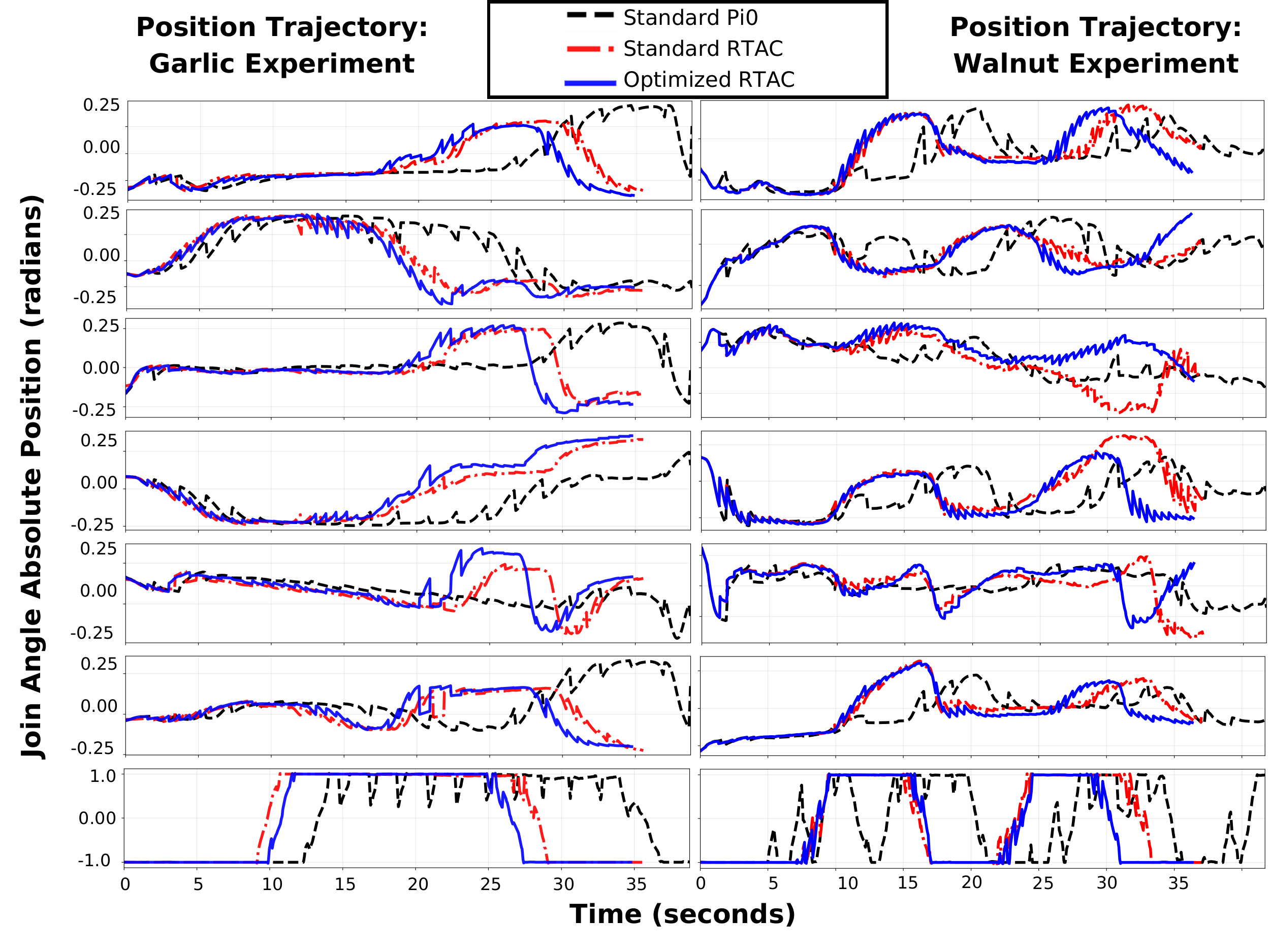}
    \caption{Positional tracking across the FR5's 6-DOF arm and the 1-DOF Jodell gripper for the garlic (left) and walnut (right) tasks. We compare synchronous standard $\pi_0$ (Black, Dashed), standard RTAC (Red, Dash-Dot), and our optimized RTAC (Blue, Solid). In order from top to bottom, the subgraphs represent Joints 1--6 of the robotic arm, with the bottommost subgraph being the Jodell gripper. It is important to note that a value of 1.0 means the gripper is closed, and a value of -1.0 means the gripper is fully open. }
    \label{fig:trajectory_divergence}   
\end{figure*}


To evaluate our proposed optimized RTAC, we design two agricultural manipulation tasks: grasping a single garlic bulb from an elevated surface and placing it into a bowl, and sequentially grasping two walnuts one at a time and depositing them into the same bowl. To provide a comprehensive comparison, each task is evaluated under three distinct configurations: i) a standard $\pi_0$ setup operating without RTAC; ii) a standard RTAC implementation derived directly from the original algorithmic pseudocode \cite{black2025real}; and iii) our proposed optimized RTAC implementation. 






\subsection{Qualitative Results}

The first task involves grasping a single garlic bulb from an elevated surface and transferring it into a bowl. Figure~\ref{fig:garlic_task} illustrates an execution of the garlic manipulation task from the perspective of the stand-mounted side camera. The execution sequence consists mainly of four distinct stages. The sequence begins with the arm approaching the garlic bulb on the elevated platform (Figure~\ref{fig:garlic_task}a). The arm then positions itself to precisely grasp the garlic (Figure~\ref{fig:garlic_task}b). Once the garlic is securely grasped, the arm moves above the bowl (Figure~\ref{fig:garlic_task}c). Finally, the arm releases the garlic into the bowl (Figure~\ref{fig:garlic_task}d). This task requires the robotic system to align the gripper with the garlic’s central body ring while maintaining the robot arm parallel to the garlic’s central plane. It must approach at a precise altitude, grasping only when accurately aligned, which is critical for successful execution.

The second task involves the sequential manipulation of two walnuts, where the robot must grasp and transfer each walnut individually from an elevated surface into a bowl. The execution procedure is illustrated in Figure~\ref{fig:walnut_task}a–d. The sequence begins with the arm approaching one of the two walnuts (Figure~\ref{fig:walnut_task}a) and grasping it. The arm then positions the walnut above the bowl and deposits it (Figure~\ref{fig:walnut_task}b). Afterwards, the robot approaches the remaining walnut and grasps it (Figure~\ref{fig:walnut_task}c), before finally releasing the second walnut into the bowl (Figure~\ref{fig:walnut_task}d).

All three algorithms, i.e., standard $\pi_0$, standard RTAC, and optimized RTAC, were evaluated on the same two tasks (Figures~\ref{fig:garlic_task} and \ref{fig:walnut_task}). Figure~\ref{fig:trajectory_divergence} shows the resulting joint angle trajectories versus runtime. The optimized RTAC completes action chunks more rapidly, as indicated by the earlier termination of the trajectory (blue), compared to both the standard $\pi_0$ (black) and standard RTAC (red).
Additionally, action chunks generated by the optimized RTAC maintain a level of smoothness comparable to that of standard RTAC while achieving faster execution. This improvement is most evident in the gripper’s angular position (bottom-most subgraph in Figure \ref{fig:trajectory_divergence}) across both experiments. 


\subsection{Quantitative Results}


We quantify the runtime performance of the three algorithms for the garlic and walnut tasks in Tables 1 and 2, respectively. All results are reported in seconds, where lower values indicate improved execution efficiency.

\begin{table*}[!t]
    \centering
    \begin{minipage}{0.48\textwidth}
        \centering
        \caption{Policy Performance Comparison for Garlic Manipulation}
        \label{tab:speedup}
        \begin{tabular}{lcc}
        \toprule
         & \textbf{Task Completion (s)} & \textbf{Episode Completion (s)} \\
        \midrule
        \textbf{Standard Pi0 \cite{black2024pi_0}} & 36.818 & 38.800 \\
        \textbf{Standard RTAC \cite{black2025real}} & 29.152 & 35.548 \\
        \midrule
        \textbf{Optimized RTAC} & \textbf{27.409} & \textbf{34.764} \\
        \bottomrule
        \end{tabular}
    \end{minipage}\hfill
    \begin{minipage}{0.48\textwidth}
        \centering
        \caption{Policy Performance Comparison for Walnut Manipulation}
        \label{tab:walnut_speedup}
        \begin{tabular}{lcc}
        \toprule
         & \textbf{Task Completion (s)} & \textbf{Episode Completion (s)} \\
        \midrule
        \textbf{Standard Pi0 \cite{black2024pi_0}} & 35.268 & 41.711 \\
        \textbf{Standard RTAC \cite{black2025real}} & 33.323 & 37.001 \\
        \midrule
        \textbf{Optimized RTAC} & \textbf{31.070} & \textbf{36.331} \\
        \bottomrule
        \end{tabular}
    \end{minipage}
\end{table*}

Performance is evaluated using two timing metrics. The first metric is (i) task completion time, defined as the time required to complete the task from the beginning of an episode. For the first task, completion is defined as the robot successfully releasing a garlic bulb into the bowl. For the second task, completion requires the robot to consecutively pick up and place two walnuts into the bowl, with success defined by both walnuts being correctly deposited. The second metric is (ii) full episode completion time, which represents the total time required to execute an entire episode. In both tasks, an episode consists of executing a complete sequence of 1000 actions.

Overall, our proposed optimized RTAC implementation achieves the lowest execution times in both evaluation metrics. For completion of the garlic task, our optimized RTAC provides a \textbf{25.6\% speedup} compared to standard $\pi_0$ and a further \textbf{6.0\% speedup} over the standard RTAC implementation. For full episode completion, Optimized RTAC achieves a \textbf{10.4\% speedup} compared to standard $\pi_0$ and a \textbf{2.2\% improvement} over standard RTAC. 

\par For the walnut task, our optimized RTAC achieves an \textbf{11.9\% speedup} over standard $\pi_0$ and a \textbf{6.8\% speedup} over standard RTAC. For full episode completion, optimized RTAC achieves a \textbf{12.9\% speedup} compared to standard $\pi_0$ and a \textbf{1.8\% speedup} over standard RTAC.
These results show that while RTAC reduces latency via asynchronous execution, system-level threading optimizations provide additional runtime gains. By executing actions faster, the optimized RTAC completes tasks more quickly. Additionally, as action horizon and task complexity grow, the benefits of reduced synchronization and improved concurrency are expected to increase, further demonstrating its performance advantage.


\section{conclusion and future work}
In this work, we addressed the critical systems-level bottlenecks that hinder the deployment of continuous Vision-Language Action models on low-cost, high-frequency robotic hardware. By translating the theoretical Real-Time Action Chunking algorithm into a highly optimized, process-based threading architecture, we effectively bypassed Python GIL limitations and mitigated latency. Our deployment on the Fairino FR5 demonstrated a reduction in task completion time and episode completion time.

Future work will focus on expanding this architecture to multi-arm cooperative systems, studying how cross-arm latency and action-chunk synchronization can affect the continuous action generation step, and dynamically scaling the chunk horizon based on the computational load of the VLA policy in real-time.

\bibliographystyle{IEEEtran}
\bibliography{refs}

\end{document}